\documentclass[runningheads]{llncs}
\usepackage{graphicx}
\usepackage{amsmath,amssymb} 
\usepackage{color}
\usepackage{tabularx}
\usepackage{multirow}
\usepackage{subfig}
\usepackage{floatrow}
\usepackage{url}
\usepackage{cite}

\DeclareMathOperator{\is}{IS}
\DeclareMathOperator*{\mean}{\mathbb{E}}
\DeclareMathOperator{\tr}{Tr}
\newcolumntype{R}{>{\raggedleft\arraybackslash}X}
\newcolumntype{Y}{>{\centering\arraybackslash}X}
\newcommand{\norm}[1]{\left\lVert#1\right\rVert}
\begin{document}

\title{How good is my GAN?}

\titlerunning{How good is my GAN?}

\authorrunning{K.\ Shmelkov \and C.\ Schmid \and K.\ Alahari}

\author{Konstantin Shmelkov\and Cordelia Schmid \and Karteek Alahari}
\institute{Inria\thanks{Univ.\ Grenoble Alpes, Inria, CNRS, Grenoble INP, LJK, 38000 Grenoble, France. This work was supported in part by the ERC advanced grant ALLEGRO, gifts from Amazon, Facebook and Intel, and the Indo-French project EVEREST (no. 5302-1) funded by CEFIPRA.
}}

\maketitle

\begin{abstract}
Generative adversarial networks (GANs) are one of the most popular methods for
generating images today. While impressive results have been validated by visual
inspection, a number of quantitative criteria have emerged only recently. We
argue here that the existing ones are insufficient and need to be in adequation
with the task at hand. In this paper we introduce two measures based on image
classification---GAN-train and GAN-test, which approximate the recall
(diversity) and precision (quality of the image) of GANs respectively. We
evaluate a number of recent GAN approaches based on these two measures and
demonstrate a clear difference in performance. Furthermore, we observe that the
increasing difficulty of the dataset, from CIFAR10 over CIFAR100 to ImageNet,
shows an inverse correlation with the quality of the GANs, as clearly evident
from our measures.
\end{abstract}

\section{Introduction}
Generative Adversarial Networks (GANs)~\cite{gan} are deep neural net
architectures composed of a pair of competing neural networks: a generator and
a discriminator. This model is trained by alternately optimizing two objective
functions so that the generator $G$ learns to produce samples resembling real
images, and the discriminator $D$ learns to better discriminate between real
and fake data. Such a paradigm has huge potential, as it can learn to generate
any data distribution. This has been exploited with some success in several
computer vision problems, such as text-to-image~\cite{stackgan} and
image-to-image~\cite{pix2pix,cyclegan} translation,
super-resolution~\cite{srgan}, and realistic natural image
generation~\cite{karras2018progressive}.

Since the original GAN model~\cite{gan} was proposed, many variants have
appeared in the past few years, for example, to improve the quality of the
generated
images~\cite{karras2018progressive,infogan,miyato2018spectral,Denton15}, or to
stabilize the training
procedure~\cite{wgan,wgangp,began,lsgan,miyato2018spectral,energygan,fgan16}.
GANs have also been modified to generate images of a given class by
conditioning on additional information, such as the class
label~\cite{cgan,dumoulin2016adversarially,miyato2018cgans,acgan}. There are a
number of ways to do this: ranging from concatenation of label $y$ to the
generator input $\mathbf{z}$ or intermediate feature
maps~\cite{dumoulin2016adversarially,cgan}, to using conditional batch
normalization~\cite{miyato2018cgans}, and augmenting the discriminator with an
auxiliary classifier~\cite{acgan}. With several such variants being regularly
proposed in the literature, a critical question is {\it how these models can be
evaluated and compared to each other.}

\begin{figure}[t]
\begin{center}
\includegraphics[width=0.8\linewidth]{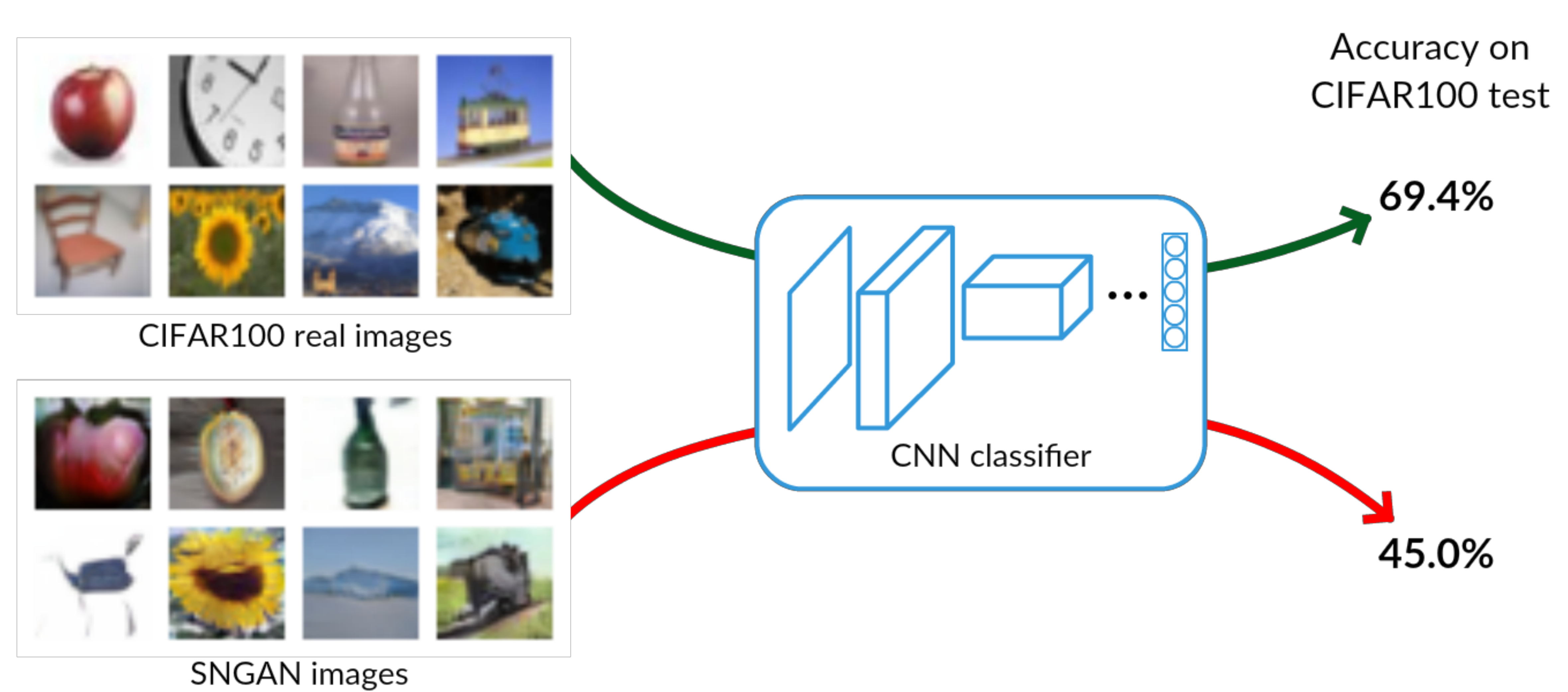}
\caption{State-of-the-art GANs, e.g., SNGAN~\cite{miyato2018spectral}, generate
realistic images, which are difficult to evaluate subjectively in comparison to
real images. Our new image classification accuracy-based measure (GAN-train is
shown here) overcomes this issue, showing a clear difference between real and
generated images.}
\label{fig:teaser}
\end{center}
\end{figure}

Evaluation and comparison of GANs, or equivalently, the images generated by
GANs, is challenging. This is in part due to the lack of an explicit likelihood
measure~\cite{theis2016note}, which is commonplace
in comparable probabilistic models~\cite{pixelcnnpp,vae}. Thus, much of the
previous work has resorted to a mere subjective visual evaluation in the case
of images synthesized by GANs. As seen from the sample images generated by a
state-of-the-art GAN~\cite{miyato2018spectral} in Figure~\ref{fig:teaser}, it
is impossible to judge their quality precisely with a subjective evaluation.
Recent work in the past two years has begun to target this challenge through
quantitative measures for evaluating
GANs~\cite{salimans2016improved,heusel2017gans,lucic2017gans,karras2018progressive}.

Inception score (IS)~\cite{salimans2016improved} and Fr\'echet Inception
distance (FID)~\cite{heusel2017gans} were suggested as ad-hoc measures
correlated with the visual quality of generated images. Inception score
measures the quality of a generated image by computing the KL-divergence
between the (logit) response produced by this image and the marginal
distribution, i.e., the average response of all the generated images, using an
Inception network~\cite{inception} trained on ImageNet. In other words,
Inception score does not compare samples with a target distribution, and is
limited to quantifying the diversity of generated samples. Fr\'echet Inception
distance compares Inception activations (responses of the penultimate layer of
the Inception network) between real and generated images. This comparison
however approximates the activations of real and generated images as Gaussian
distributions (cf.\ equation (\ref{eqn:fid})), computing their means and
covariances, which are too crude to capture subtle details. Both these measures
rely on an ImageNet-pretrained Inception network, which is far from ideal for
other datasets, such as faces and biomedical imaging. Overall, IS and FID are
useful measures to evaluate how training advances, but they guarantee no
correlation with performance on real-world tasks. As we discuss in
Section~\ref{sec:expts}, these measures are insufficient to finely separate
state-of-the-art GAN models, unlike our measures (see SNGAN vs WPGAN-GP (10M)
in Table~\ref{tab:cifar100} for example).

An alternative evaluation is to compute the distance of the generated samples
to the real data manifold in terms of precision and
recall~\cite{lucic2017gans}. Here, high precision implies that the generated
samples are close to the data manifold, and high recall shows that the
generator outputs samples that cover the manifold well. These measures remain
idealistic as they are impossible to compute on natural image data, whose
manifold is unknown. Indeed, the evaluation in~\cite{lucic2017gans} is limited
to using synthetic data composed of gray-scale triangles. Another distance
suggested for comparing GAN models is sliced Wasserstein distance
(SWD)~\cite{karras2018progressive}. SWD is an approximation of Wasserstein-1
distance between real and generated images, and is computed as the statistical
similarity between local image patches extracted from Laplacian pyramid
representations of these images. As shown in Section~\ref{sec:expts}, SWD is
less-informative than our evaluation measures.

In this paper, we propose new evaluation measures to compare class-conditional
GAN architectures with GAN-train and GAN-test scores. We rely on a neural net
architecture for image classification for both these measures. To compute
GAN-train, we train a classification network with images generated by a GAN,
and then evaluate its performance on a test set composed of real-world images.
Intuitively, this measures the difference between the learned (i.e., generated
image) and the target (i.e., real image) distributions. We can conclude that
generated images are similar to real ones if the classification network, which
learns features for discriminating images generated for different classes, can
correctly classify real images. In other words, GAN-train is akin to a recall
measure, as a good GAN-train performance shows that the generated samples are
diverse enough. However, GAN-train also requires a sufficient precision, as
otherwise the classifier will be impacted by the sample quality.  

Our second measure, GAN-test, is the accuracy of a network trained on real
images and evaluated on the generated images. This measure is similar to
precision, with a high value denoting that the generated samples are a
realistic approximation of the (unknown) distribution of natural images. In
addition to these two measures, we study the utility of images generated by
GANs for augmenting training data. This can be interpreted as a measure of the
diversity of the generated images. The utility of our evaluation approach, in
particular, when a subjective inspection is insufficient, is illustrated with
the GAN-train measure in Figure~\ref{fig:teaser}. We will discuss these
measures in detail in Section~\ref{sec:approach}.

As shown in our extensive experimental results in Section~\ref{sec:expts} and
the appendix in the supplementary material and technical report~\cite{suppmat},
these measures are much more informative to evaluate GANs, compared to all the
previous measures discussed, including cases where human studies are
inconclusive. In particular, we evaluate two state-of-the-art GAN models:
WGAN-GP~\cite{wgangp} and SNGAN~\cite{miyato2018spectral}, along with other
generative models~\cite{dcgan,pixelcnnpp} to provide baseline comparisons.
Image classification performance is evaluated on MNIST~\cite{MNIST}, CIFAR10,
CIFAR100~\cite{cifar}, and the ImageNet~\cite{imagenet} datasets. Experimental
results show that the quality of GAN images decreases significantly as the
complexity of the dataset increases.

\section{Related work}
We present existing quantitative measures to evaluate GANs: scores based on an
Inception network, i.e., IS and FID, a Wasserstein-based distance metric,
precision and recall scores, and a technique built with data augmentation.

\subsection{Inception score}%
One of the most common ways to evaluate GANs is the Inception
score~\cite{salimans2016improved}. It uses an Inception
network~\cite{inception} pre-trained on ImageNet to compute logits of generated
images. The score is given by:
\begin{equation}
  \is (G) = \exp \left(\mean_{\mathbf{x}\backsim p_g} [D_{\text{KL}}(p(y|\mathbf{x}) \parallel p(y))] \right),
\end{equation}
where $\mathbf{x}$ is a generated image sampled from the learned generator
distribution $p_g$, $\mean$ is the expectation over the set of generated
images, $D_{\text{KL}}$ is the KL-divergence between the conditional class
distribution $p(y|\mathbf{x})$ (for label $y$, according to the Inception
network) and the marginal class distribution \mbox{$p(y) =
\mean\limits_{\mathbf{x} \backsim p_g} [p(y | \mathbf{x})$}]. By definition,
Inception score does not consider real images at all, and so cannot measure how
well the generator approximates the real distribution. This score is limited to
measuring only the diversity of generated images. Some of its other
limitations, as noted in~\cite{barratt2018note}, are: high sensitivity to small
changes in weights of the Inception network, and large variance of scores.

\subsection{Fr\'{e}chet Inception distance}%
The recently proposed Fr\'{e}chet Inception distance
(FID)~\cite{heusel2017gans} compares the distributions of Inception embeddings
(activations from the penultimate layer of the Inception network) of real
($p_r(\mathbf{x})$) and generated ($p_g(\mathbf{x})$) images. Both these
distributions as modeled as multi-dimensional Gaussians parameterized by their
respective mean and covariance. The distance measure is defined between the two
Gaussian distributions as:
\begin{equation}
  d^2((\mathbf{m}_r, \mathbf{C}_r), (\mathbf{m}_g, \mathbf{C}_g)) =
  \norm{\mathbf{m}_r - \mathbf{m}_g}^2 + \tr (\mathbf{C}_r + \mathbf{C}_g - 2(\mathbf{C}_r \mathbf{C}_g)^\frac{1}{2}),
\label{eqn:fid}
\end{equation}
where $(\mathbf{m}_r, \mathbf{C}_r)$, $(\mathbf{m}_g, \mathbf{C}_g)$ denote the
mean and covariance of the real and generated image distributions respectively.
FID is inversely correlated with Inception score, and suffers from the same
issues discussed earlier.

The two Inception-based measures cannot separate image quality from image
diversity. For example, low IS or FID values can be due to the generated images
being either not realistic (low image quality) or too similar to each other
(low diversity), with no way to analyze the cause. In contrast, our measures
can distinguish when generated images become less diverse from worse image
quality.

\subsection{Other evaluation measures}%
Sliced Wasserstein distance (SWD)~\cite{karras2018progressive} was used to
evaluate high-resolution GANs. It is a multi-scale statistical similarity
computed on local image patches extracted from the Laplacian pyramid
representation of real and generated images. A total of 128 $7\times 7$ local
patches for each level of the Laplacian pyramid are extracted per image. While
SWD is an efficient approximation, using randomized
projections~\cite{rabin2011wasserstein}, of the Wasserstein-1 distance between
the real and generated images, its utility is limited when comparing a variety
of GAN models, with not all of them producing high-resolution images (see our
evaluation in Section~\ref{sec:expts}).

Precision and recall measures were introduced~\cite{lucic2017gans} in the
context of GANs, by constructing a synthetic data manifold. This makes it
possible to compute the distance of an image sample (generated or real) to the
manifold, by finding its distance to the closest point from the manifold. In
this synthetic setup, precision is defined as the fraction of the generated
samples whose distance to the manifold is below a certain threshold. Recall, on
the other hand, is computed by considering a set of test samples. First, the
latent representation $\mathbf{\tilde{z}}$ of each test sample $\mathbf{x}$ is
estimated, through gradient descent, by inverting the generator $G$. Recall is
then given by the fraction of test samples whose L2-distance to
$G(\mathbf{\tilde{z}})$ is below the threshold. High recall is equivalent to
the GAN capturing most of the manifold, and high precision implies that the
generated samples are close to the manifold. Although these measures bring the
flavor of techniques used widely to evaluate discriminative models to GANs,
they are impractical for real images as the data manifold is unknown, and their
use is limited to evaluations on synthetic data~\cite{lucic2017gans}.

\subsection{Data augmentation}%
Augmenting training data is an important component of learning neural networks.
This can be achieved by increasing the size of the training set~\cite{alexnet}
or incorporating augmentation directly in the latent space~\cite{wang2018low}.
A popular technique is to increase the size of the training set with minor
transformations of data, which has resulted in a performance boost, e.g., for
image classification~\cite{alexnet}. GANs provide a natural way to augment
training data with the generated samples. Indeed, GANs have been used to train
classification networks in a semi-supervised
fashion~\cite{dai2017good,tran2017bayesian} or to facilitate domain
adaptation~\cite{bousmalis2017unsupervised}. Modern GANs generate images
realistic enough to improve performance in applications, such as, biomedical
imaging~\cite{frid2018synthetic,calimeri2017biomedical}, person
re-identification~\cite{zhong2018camera} and image
enhancement~\cite{yun2018predicting}. They can also be used to refine training
sets composed of synthetic images for applications such as eye gaze and hand
pose estimation~\cite{shrivastava2017learning}. GANs are also used to learn
complex 3D distributions and replace computationally intensive simulations in
physics~\cite{paganini2018accelerating,mosser2017reconstruction} and
neuroscience~\cite{molano2018synthesizing}. Ideally, GANs should be able to
recreate the training set with different variations. This can be used to
compress datasets for learning incrementally, without suffering from
catastrophic forgetting as new classes are added~\cite{shin2017continual}. We
will study the utility of GANs for training image classification networks with
data augmentation (see Section~\ref{sec:augment}), and analyze it as an
evaluation measure.

\begin{figure}[t]
\floatbox[{\capbeside\thisfloatsetup{capbesideposition={right,top},capbesidewidth=6cm}}]{figure}[\FBwidth]
{\caption{Illustration of GAN-train and GAN-test. GAN-train learns a classifier
on GAN generated images and measures the performance on real test images. This
evaluates the diversity and realism of GAN images. GAN-test learns a classifier
on real images and evaluates it on GAN images. This measures how realistic GAN
images are.}\label{fig:scheme}}
{\includegraphics[width=\linewidth]{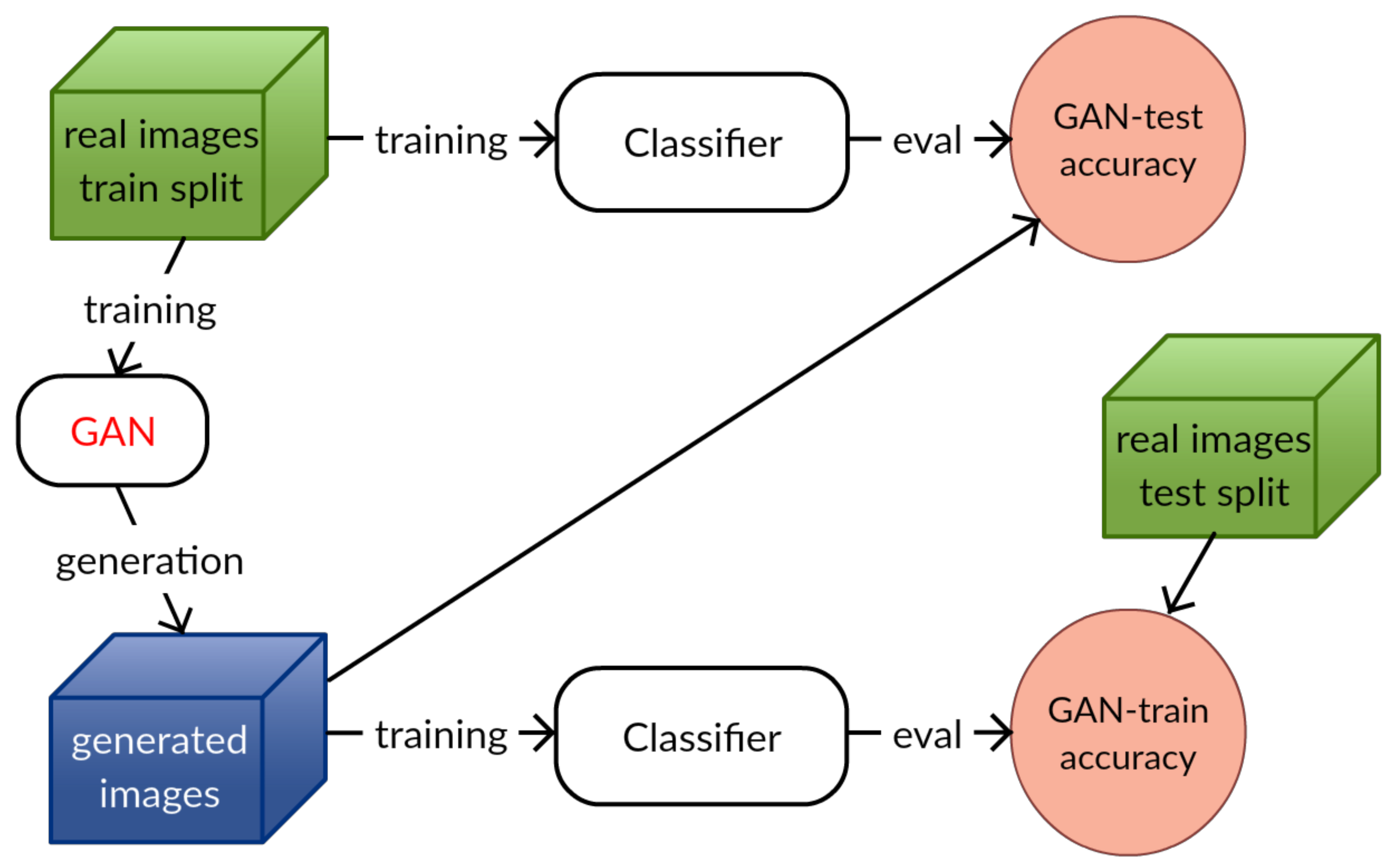}}
\end{figure}

In summary, evaluation of generative models is not a easy
task~\cite{theis2016note}, especially for models like GANs. We bring a new
dimension to this problem with our GAN-train and GAN-test performance-based
measures, and show through our extensive analysis that they are complementary
to all the above schemes.

\section{GAN-train and GAN-test}\label{sec:approach}%
An important characteristic of a conditional GAN model is that generated images
should not only be realistic, but also recognizable as coming from a given
class. An optimal GAN that perfectly captures the target distribution can
generate a new set of images $S_g$, which are indistinguishable from the
original training set $S_t$. Assuming both these sets have the same size, a
classifier trained on either of them should produce roughly the same validation
accuracy. This is indeed true when the dataset is simple enough, for example,
MNIST~\cite{shin2017continual} (see also Section~\ref{sec:evaluation}).
Motivated by this optimal GAN characteristic, we devise two scores to evaluate
GANs, as illustrated in Figure~\ref{fig:scheme}.

GAN-train is the accuracy of a classifier trained on $S_g$ and tested on a
validation set of real images $S_v$. When a GAN is not perfect, GAN-train
accuracy will be lower than the typical validation accuracy of the classifier
trained on $S_t$. It can happen due to many reasons, e.g., (i) mode dropping
reduces the diversity of $S_g$ in comparison to $S_t$, (ii) generated samples
are not realistic enough to make the classifier learn relevant features, (iii)
GANs can mix-up classes and confuse the classifier. Unfortunately, GAN failures
are difficult to diagnose.
When GAN-train accuracy is
close to validation accuracy, it means that GAN images are high quality and as
diverse as the training set. As we will show in Section~\ref{sec:diversity},
diversity varies with the number of generated images. We will analyze this with
the evaluation discussed at the end of this section.

GAN-test is the accuracy of a classifier trained on the original training set
$S_t$, but tested on $S_g$. If a GAN learns well, this turns out be an easy
task because both the sets have the same distribution. Ideally, GAN-test should
be close to the validation accuracy. If it significantly higher, it means that
the GAN overfits, and simply memorizes the training set. On the contrary, if it
is significantly lower, the GAN does not capture the target distribution well
and the image quality is poor. Note that this measure does not capture the
diversity of samples because a model that memorizes exactly one training image
perfectly will score very well. GAN-test accuracy is related to the precision
score in~\cite{lucic2017gans}, quantifying how close generated images are to a
data manifold.

To provide an insight into the diversity of GAN-generated images, we measure
GAN-train accuracy with generated sets of different sizes, and compare it with
the validation accuracy of a classifier trained on real data of the
corresponding size. If all the generated images were perfect, the size of $S_g$
where GAN-train is equal to validation accuracy with the reduced-size training
set, would be a good estimation of the number of distinct images in $S_g$. In
practice, we observe that GAN-train accuracy saturates with a certain number of
GAN-generated samples (see Figures~\ref{fig:cifar_gen_data}(a)
and~\ref{fig:cifar_gen_data}(b) discussed in Section~\ref{sec:diversity}). This
is a measure of the diversity of a GAN, similar to recall
from~\cite{lucic2017gans}, measuring the fraction of the data manifold covered
by a GAN.

\section{Datasets and Methods}
\noindent {\bf Datasets.} For comparing the different GAN methods and
PixelCNN++, we use several image classification datasets with an increasing
number of labels: MNIST~\cite{MNIST}, CIFAR10~\cite{cifar},
CIFAR100~\cite{cifar} and ImageNet1k~\cite{imagenet}.  CIFAR10 and CIFAR100
both have 50k $32\times32$ RGB images in the training set, and 10k images in
the validation set. CIFAR10 has 10 classes while CIFAR100 has 100 classes.
ImageNet1k has 1000 classes with 1.3M training and 50k validation images. We
downsample the original ImageNet images to two resolutions in our experiments,
namely $64\times64$ and $128\times128$. MNIST has 10 classes of $28\times28$
grayscale images, with 60k samples for training and 10k for validation.

We exclude the CIFAR10/CIFAR100/ImageNet1k validation images from GAN training
to enable the evaluation of test accuracy. This is not done in a number of GAN
papers and may explain minor differences in IS and FID scores compared to the
ones reported in these papers.

\subsection{Evaluated methods}
Among the plethora of GAN models in literature, it is difficult to choose the
best one, especially since appropriate hyperparameter fine-tuning appears to
bring all major GANs within a very close performance range, as noted in a
study~\cite{lucic2017gans}. We choose to perform our analysis on Wasserstein
GAN (WGAN-GP), one of the most widely-accepted models in literature at the
moment, and SNGAN, a very recent model showing state-of-the-art image
generation results on ImageNet. Additionally, we include two baseline
generative models, DCGAN~\cite{dcgan} and PixelCNN++~\cite{pixelcnnpp}.
We summarize all the models included in our experimental analysis below, and
present implementation details in the appendix~\cite{suppmat}.

\noindent{\bf Wasserstein GAN.} WGAN~\cite{wgan} replaces the discriminator separating real and generated images with a critic estimating Wasserstein-1 (i.e., earth-mover's) distance between their corresponding distributions.  
The success of WGANs in comparison to the classical GAN model~\cite{gan}
can be attributed to two reasons. Firstly, the optimization of the generator is
easier because the gradient of the critic function is better behaved than its
GAN equivalent. Secondly, empirical observations show that the WGAN value
function better correlates with the quality of the samples than
GANs~\cite{wgan}. 

In order to estimate the Wasserstein-1 distance between the real and generated
image distributions, the critic must be a K-Lipschitz function. The original
paper~\cite{wgan} proposed to constrain the critic through weight clipping to
satisfy this Lipschitz requirement. This, however, can lead to unstable
training or generate poor samples~\cite{wgangp}. An alternative to clipping
weights is the use of a gradient penalty as a regularizer to enforce the
Lipschitz constraint. In particular, we penalize the norm of the gradient of
the critic function with respect to its input. This has demonstrated stable
training of several GAN architectures~\cite{wgangp}.

We use the gradient penalty variant of WGAN, conditioned on data in our
experiments, and refer to it as WGAN-GP in the rest of the paper. Label
conditioning is an effective way to use labels available in image
classification training data~\cite{acgan}. Following ACGAN~\cite{acgan}, we
concatenate the noise input $\mathbf{z}$ with the class label in the generator,
and modify the discriminator to produce probability distributions over the
sources as well as the labels. 

\noindent {\bf SNGAN.} Variants have also analyzed other issues related to
training GANs, such as the impact of the performance control of the
discriminator on training the generator. Generators often fail to learn the
multimodal structure of the target distribution due to unstable training of the
discriminator, particularly in high-dimensional
spaces~\cite{miyato2018spectral}. More dramatically, generators cease to learn
when the supports of the real and the generated image distributions are
disjoint~\cite{arjovsky2017towards}. This occurs since the discriminator
quickly learns to distinguish these distributions, resulting in the gradients
of the discriminator function, with respect to the input, becoming zeros, and
thus failing to update the generator model any further.

SNGAN~\cite{miyato2018spectral} introduces spectral normalization to stabilize
training the discriminator. This is achieved by normalizing each layer of the
discriminator (i.e., the learnt weights) with the spectral norm of the weight
matrix, which is its largest singular value. Miyato {\it et al.\
}\cite{miyato2018spectral} showed that this regularization outperforms other
alternatives, including gradient penalty, and in particular, achieves
state-of-the-art image synthesis results on ImageNet. We use the
class-conditioned version of SNGAN~\cite{miyato2018cgans} in our evaluation.
Here, SNGAN is conditioned with projection in the discriminator network, and
conditional batch normalization~\cite{cbn} in the generator network.

\noindent {\bf DCGAN.} Deep convolutional GANs (DCGANs) is a class of architecture
that was proposed to leverage the benefits of supervised learning with CNNs as
well as the unsupervised learning of GAN models~\cite{dcgan}. The main
principles behind DCGANs are using only convolutional layers and batch
normalization for the generator and discriminator networks.
Several instantiations of DCGAN are possible with these broad
guidelines, and in fact, many do exist in
literature~\cite{acgan,wgangp,miyato2018spectral}. We use the class-conditioned
variant presented in~\cite{acgan} for our analysis.

\noindent {\bf PixelCNN++.} The original PixelCNN~\cite{pixelcnn} belongs to a
class of generative models with tractable likelihood. It is a deep neural net
which predicts pixels sequentially along both the spatial dimensions. The
spatial dependencies among pixels are captured with a fully convolutional
network using masked convolutions. PixelCNN++ proposes improvements to this
model in terms of regularization, modified network connections
and more efficient training~\cite{pixelcnnpp}.

\section{Experiments}\label{sec:expts}%
\subsection{Implementation details of evaluation measures}
We compute Inception score with the WGAN-GP code\cite{urlwgan} corrected for
the 1008 classes problem~\cite{barratt2018note}. The mean value of this score
computed 10 times on 5k splits is reported in all our evaluations, following
standard protocol.

We found that there are two variants for computing FID. The first one is the
original implementation\cite{urlfid} from the authors~\cite{heusel2017gans},
where all the real images and at least 10k generated images are used. The
second one is from the SNGAN~\cite{miyato2018spectral} implementation, where 5k
generated images are compared to 5k real images. Estimation of the covariance
matrix is also different in both these cases. Hence, we include these two
versions of FID in the paper to facilitate comparison in the future. The
original implementation is referred to as FID, while our
implementation~\cite{ourcode} of the 5k version is denoted as FID-5K.
Implementation of SWD is taken from the official NVIDIA
repository\cite{urlswd}.

\subsection{Generative model evaluation}\label{sec:evaluation}
\noindent{\bf MNIST.}
We validate our claim (from Section~\ref{sec:approach}) that a GAN can
perfectly reproduce a simple dataset on MNIST. A four-layer convnet classifier
trained on real MNIST data achieves 99.3\% accuracy on the test set. In
contrast, images generated with SNGAN achieve a GAN-train accuracy of 99.0\%
and GAN-test accuracy of 99.2\%, highlighting their high image quality as well
as diversity.

\noindent{\bf CIFAR10.}
Table~\ref{tab:cifar10} shows a comparison of state-of-the-art GAN models on
CIFAR10. We observe that the relative ranking of models is consistent across
different metrics: FID, GAN-train and GAN-test accuracies. Both GAN-train and
GAN-test are quite high for SNGAN and WGAN-GP (10M). This implies that both the
image quality and the diversity are good, but are still lower than that of real
images (92.8 in the first row). Note that PixelCNN++ has low diversity because
GAN-test is much higher than GAN-train in this case. This is in line with its
relatively poor Inception score and FID (as shown in~\cite{lucic2017gans}
FID is quite sensitive to mode dropping).

\begin{table}[tbp]
\centering%
\begin{tabularx}{1.0\textwidth}{l | R | R | R | R | R | R | R}
  model          & IS   & FID-5K & FID & GAN-train & GAN-test & SWD 16 & SWD 32\\
      \hline
  real images    & 11.33& 9.4    & 2.1    & 92.8 & -    & 2.8 & 2.0 \\
  SNGAN          & 8.43 & 18.8   & 11.8   & 82.2 & 87.3 & 3.9 & 24.4 \\
  WGAN-GP (10M)  & 8.21 & 21.5   & 14.1   & 79.5 & 85.0 & 3.8 & 6.2 \\
  WGAN-GP (2.5M) & 8.29 & 22.1   & 15.0   & 76.1 & 80.7 & 3.4 & 6.9 \\
  DCGAN          & 6.69 & 42.5   & 35.6   & 65.0 & 58.2 & 6.5 & 24.7 \\
  PixelCNN++     & 5.36 & 121.3  & 119.5  & 34.0 & 47.1 & 14.9& 56.6 \\
\end{tabularx}
\caption{CIFAR10 experiments. IS: higher is better. FID and SWD: lower is
better. SWD values here are multiplied by $10^3$ for better readability.
GAN-train and GAN-test are accuracies given as percentage (higher is better).}
\label{tab:cifar10}
\end{table}%
\begin{figure}[tb]
\floatbox[{\capbeside\thisfloatsetup{capbesideposition={right,top},capbesidewidth=6.5cm}}]{figure}[\FBwidth]
{\caption{First column: SNGAN-generated images. Other columns: 5 images from
  CIFAR10 ``train'' closest to GAN image from the first column
  in feature space of baseline CIFAR10 classifier.}\label{fig:knn}}
{\includegraphics[width=0.9\linewidth]{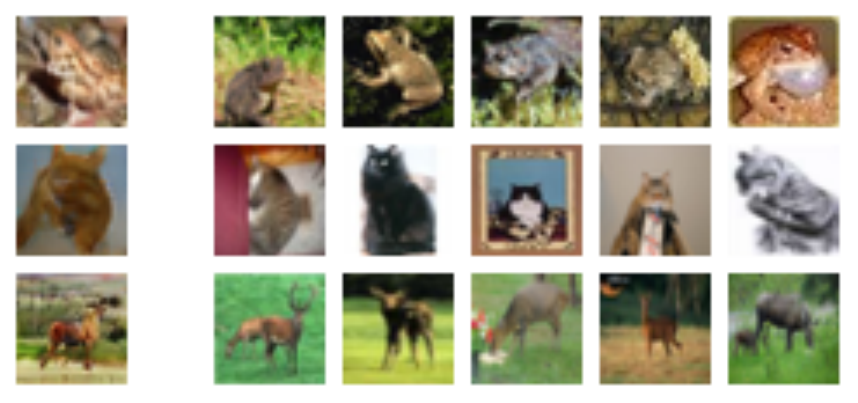}}
\end{figure}%
Note that SWD does not correlate well with other metrics: it is consistently
smaller for WGAN-GP (especially SWD 32). We hypothesize that this is because
SWD approximates the Wasserstein-1 distance between patches of real and
generated images, which is related to the optimization objective of Wasserstein
GANs, but not other models (e.g., SNGAN). This suggests that SWD is unsuitable
to compare WGAN and other GAN losses. It is also worth noting that WGAN-GP
(10M) shows only a small improvement over WGAN-GP (2.5M) despite a four-fold
increase in the number of parameters. In Figure~\ref{fig:knn} we show
SNGAN-generated images on CIFAR10 and their nearest neighbors from the training
set in the feature space of the classifier we use to compute the GAN-test
measure. Note that SNGAN consistently finds images of the same class as a
generated image, which are close to an image from the training set.

To highlight the complementarity of GAN-train and GAN-test, we emulate a simple
model by subsampling/corrupting the CIFAR10 training set, in the spirit of
\cite{heusel2017gans}. GAN-train/test now corresponds to training/testing the
classifier on modified data. We observe that GAN-test is insensitive to
subsampling unlike GAN-train (where it is equivalent to training a classifier
on a smaller split). Salt and pepper noise, ranging from 1\% to 20\% of
replaced pixels per image, barely affects GAN-train, but degrades GAN-test
significantly (from 82\% to 15\%).

Through this experiment on modified data, we also observe that FID is
insufficient to distinguish between the impact of image diversity and quality.
For example, FID between CIFAR10 train set and train set with Gaussian noise
($\sigma=5$) is 27.1, while FID between train set and its random 5k subset with
the same noise is 29.6. This difference may be due to lack of diversity or
quality or both. GAN-test, which measures the quality of images, is identical
(95\%) in both these cases. GAN-train, on the other hand, drops from 91\% to
80\%, showing that the 5k train set lacks diversity. Together, our measures,
address one of the main drawbacks of FID.

\noindent{\bf CIFAR100.}
Our results on CIFAR100 are summarized in Table~\ref{tab:cifar100}. It is a
more challenging dataset than CIFAR10, mainly due to the larger number of
classes and fewer images per class; as evident from the accuracy of a convnet
for classification trained with real images: 92.8 vs 69.4 for CIFAR10 and
CIFAR100 respectively. SNGAN and WGAN-GP (10M) produce similar IS and FID, but
very different GAN-train and GAN-test accuracies. This makes it easier to
conclude that SNGAN has better image quality and diversity than WGAN-GP (10M).
It is also interesting to note that WGAN-GP (10M) is superior to WGAN-GP (2.5M)
in all the metrics, except SWD. WGAN-GP (2.5M) achieves reasonable IS and FID,
but the quality of the generated samples is very low, as evidenced by GAN-test
accuracy. SWD follows the same pattern as in the CIFAR10 case: WGAN-GP shows a
better performance than others in this measure, which is not consistent with
its relatively poor image quality. PixelCNN++ exhibits an interesting behavior,
with high GAN-test accuracy, but very low GAN-train accuracy, showing that it
can generate images of acceptable quality, but they lack diversity. A high FID
in this case also hints at significant mode dropping. We also analyze the
quality of the generated images with t-SNE~\cite{tsne} in the
appendix~\cite{suppmat}.

\noindent{\it Random forests.}
We verify if our findings depend on the type of classifier by using random
forests~\cite{randomforests,scikit-learn} instead of CNN for classification.
This results in GAN-train, GAN-test scores of 15.2\%, 19.5\% for SNGAN, 10.9\%,
16.6\% for WGAN-GP~(10M), 3.7\%, 4.8\% for WGAN-GP~(2.5M), and 3.2\%, 3.0\% for
DCGAN respectively. Note that the relative ranking of these GANs remains
identical for random forests and CNNs.

\noindent{\it Human study.} We designed a human study with the goal of finding
which of the measures (if any) is better aligned with human judgement. The
subjects were asked to choose the more realistic image from two samples
generated for a particular class of CIFAR100. Five subjects evaluated SNGAN vs
one of the following: DCGAN, WGAN-GP~(2.5M), WGAN-GP~(10M) in three separate
tests. They made 100 comparisons of randomly generated image pairs for each
test, i.e., 1500 trials in total. All of them found the task challenging, in
particular for both WGAN-GP tests.

We use Student's t-test for statistical analysis of these results. In SNGAN vs
DCGAN, subjects chose SNGAN 368 out of 500 trials, in SNGAN vs WGAN-GP~(2.5M),
subjects preferred SNGAN 274 out of 500 trials, and in SNGAN vs WGAN-GP~(10M),
SNGAN was preferred 230 out of 500. The preference of SNGAN over DCGAN is
statistically significant ($p < 10^{-7}$), while the preference over
WGAN-GP~(2.5M) or WGAN-GP~(10M) is insignificant ($p = 0.28$ and $p = 0.37$
correspondingly). We conclude that the quality of images generated needs to be
significantly different, as in the case of SNGAN vs DCGAN, for human studies to
be conclusive. They are insufficient to pick out the subtle, but
performance-critical, differences, unlike our measures.

\begin{table}[tbp]
\centering%
\begin{tabularx}{1.0\textwidth}{l | R | R | R | R | R | R | R}
    model          & IS   & FID-5K & FID & GAN-train & GAN-test & SWD 16 & SWD 32\\
        \hline
    real images    & 14.9 & 10.8 & 2.4      & 69.4 &    - & 2.7 & 2.0 \\
    SNGAN          & 9.30 & 23.8 & 15.6     & 45.0 & 59.4 & 4.0 & 15.6 \\
    WGAN-GP (10M)  & 9.10 & 23.5 & 15.6     & 26.7 & 40.4 & 6.0 & 9.1 \\
    WGAN-GP (2.5M) & 8.22 & 28.8 & 20.6     & 5.4  &  4.3 & 3.7 & 7.7 \\
    DCGAN          & 6.20 & 49.7 & 41.8     & 3.5  &  2.4 & 9.9 & 20.8 \\
    PixelCNN++     & 6.27 & 143.4& 141.9    & 4.8  & 27.5 & 8.5 & 25.9 \\
\end{tabularx}
\caption{CIFAR100 experiments. Refer to the caption of Table~\ref{tab:cifar10}
for details.}
\label{tab:cifar100}
\end{table}
\begin{table}[tb]
\centering%
\begin{tabularx}{1.0\textwidth}{c | l | R | R | R | R | R | R| R }
  res & model & IS & FID-5K & FID & GAN-train top-1 & GAN-train top-5 & GAN-test
  top-1 & GAN-test top-5 \\
      \hline
  \multirow{3}{*}{64px} & real images  & 63.8 & 15.6 & 2.9 & 55.0 & 78.8 & - & -\\
                        & SNGAN        & 12.3 & 44.5 & 34.4 & 3 & 8.4 & 12.9 & 28.9\\
                        & WGAN-GP      & 11.3 & 46.7 & 35.8 & 0.1 & 0.7 & 0.1 & 0.5\\
      \hline
  \multirow{3}{*}{128px} & real images & 203.2 & 17.4 & 3.0 & 59.1 & 81.9 & - & - \\
                         & SNGAN*      & 35.3  & 44.9 & 33.2 & 9.3  & 21.9 & 39.5 & 63.4\\
                         & WGAN-GP     & 11.6  & 91.6 & 79.5 & 0.1  & 0.5 & 0.1 & 0.5 \\
\end{tabularx}
\caption{ImageNet experiments. SNGAN* refers to the model provided
by~\cite{miyato2018spectral}, trained for 850k iterations. Refer to the caption
of Table~\ref{tab:cifar10} for details.}
\label{tab:imagenet_exps}
\end{table}
\noindent{\bf ImageNet.}
On this dataset, which is one of the more challenging ones for image
synthesis~\cite{miyato2018spectral}, we analyzed the performance of the two
best GAN models based on our CIFAR experiments, i.e., SNGAN and WGAN-GP. As
shown in Table~\ref{tab:imagenet_exps}, SNGAN achieves a reasonable GAN-train
accuracy and a relatively high GAN-test accuracy at $128\times128$ resolution.
This suggests that SNGAN generated images have good quality, but their
diversity is much lower than the original data. This may be partly due to the
size of the generator (150Mb) being significantly smaller in comparison to
ImageNet training data (64Gb for $128\times128$). Despite this difference in
size, it achieves GAN-train accuracy of 9.3\% and 21.9\% for top-1 and top-5
classification results respectively. In comparison, the performance of WGAN-GP
is dramatically poorer; see last row for each resolution in the table.

\begin{figure}[htb]%
    \centering%
    \subfloat{{\includegraphics[width=0.45\linewidth]{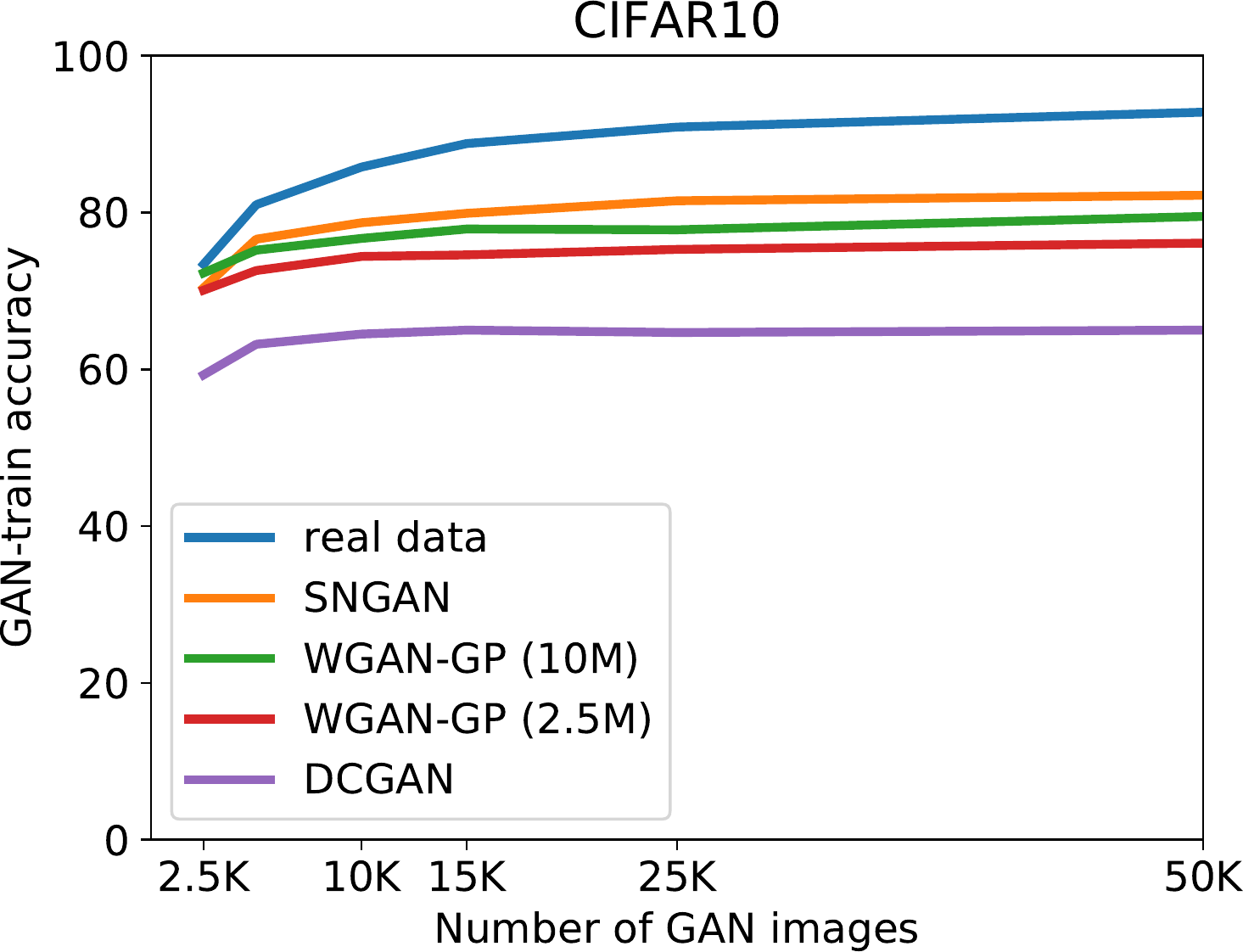} }\label{fig:cifar10_gen_data}}%
    \subfloat{{\includegraphics[width=0.45\linewidth]{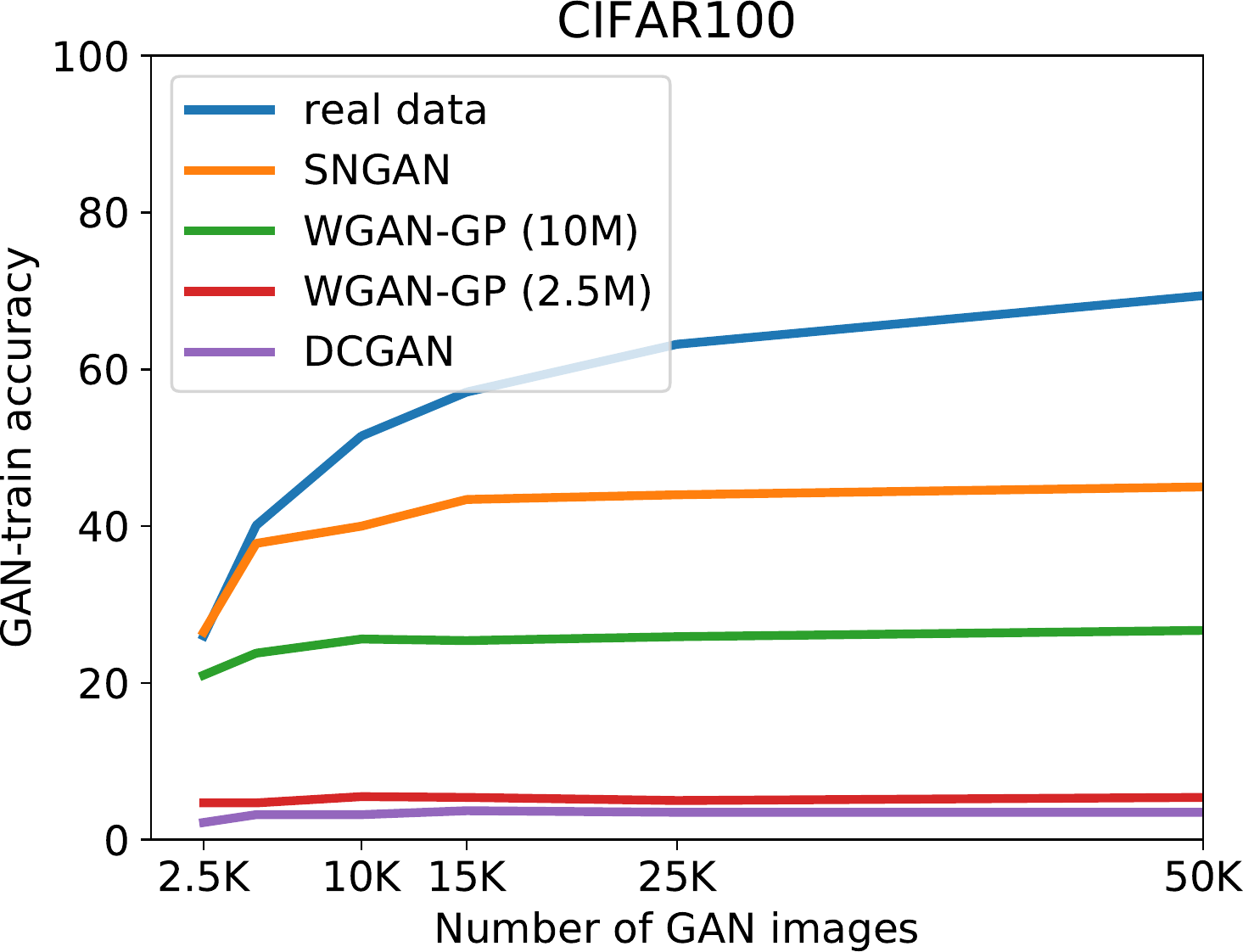} }\label{fig:cifar100_gen_data}}%
    \caption{The effect of varying the size of the generated image set on
GAN-train accuracy. For comparison, we also show the result (in blue) of
varying the size of the real image training dataset. (Best viewed in pdf.)}%
    \label{fig:cifar_gen_data}%
\end{figure}

In the case of images generated at $64\times64$ resolution, GAN-train and
GAN-test accuracies with SNGAN are lower than their $128\times128$
counterparts. GAN-test accuracy is over four times better than GAN-train,
showing that the generated images lack in diversity. It is interesting to note
that WGAN-GP produces Inception score and FID very similar to SNGAN, but its
images are insufficient to train a reasonable classifier and to be recognized
by an ImageNet classifier, as shown by the very low GAN-train and GAN-test
scores.

\subsection{GAN image diversity}\label{sec:diversity}
We further analyze the diversity of the generated images by evaluating
GAN-train accuracy with varying amounts of generated data. A model with low
diversity generates redundant samples, and increasing the quantity of data
generated in this case does not result in better GAN-train accuracy. In
contrast, generating more samples from a model with high diversity produces a
better GAN-train score. We show this analysis in
Figure~\ref{fig:cifar_gen_data}, where GAN-train accuracy is plotted with
respect to the size of the generated training set on CIFAR10 and CIFAR100.

\begin{figure}[htb]%
    \centering%
    \subfloat{{\includegraphics[width=0.45\linewidth]{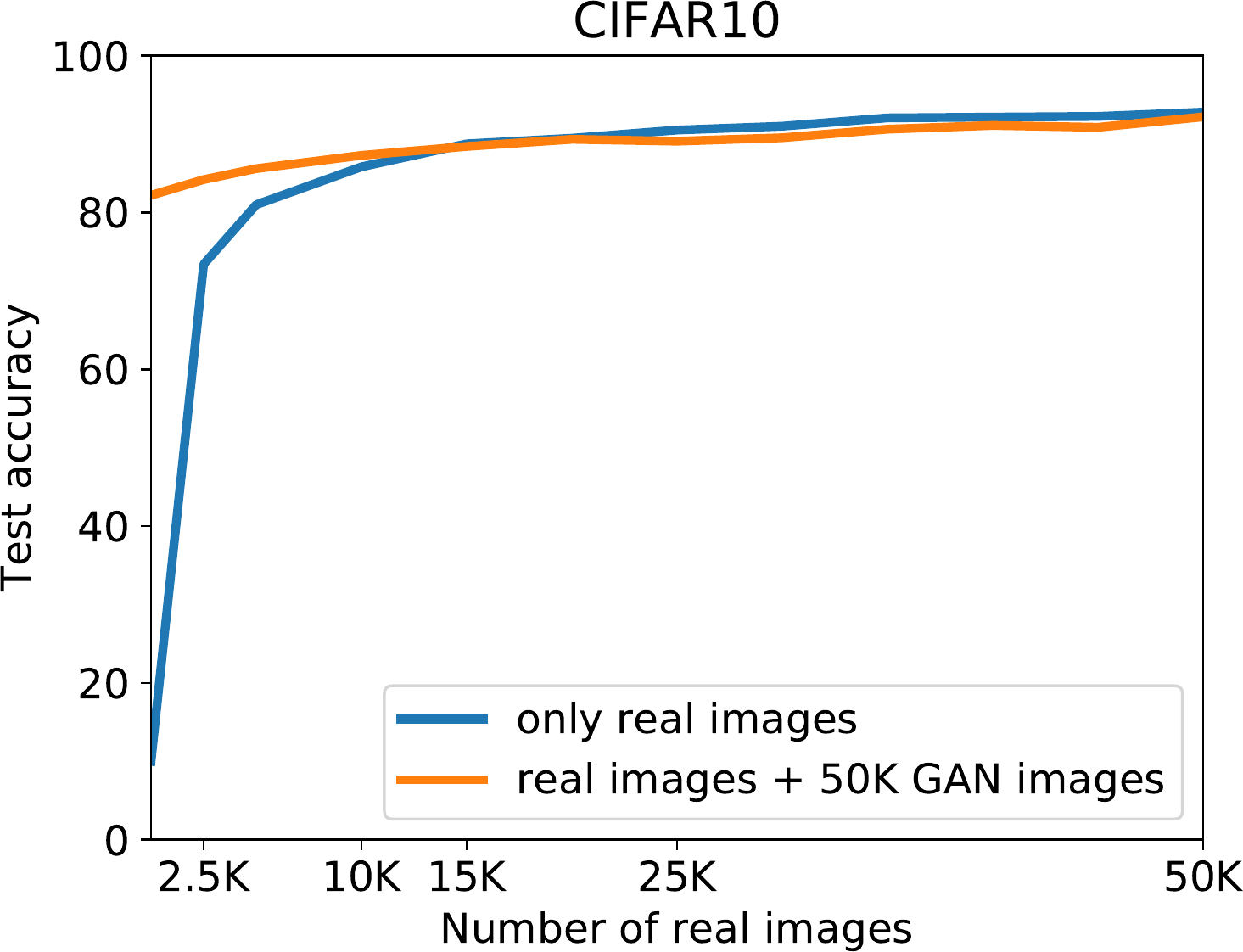} }\label{fig:cifar10_real_data}}%
    \subfloat{{\includegraphics[width=0.45\linewidth]{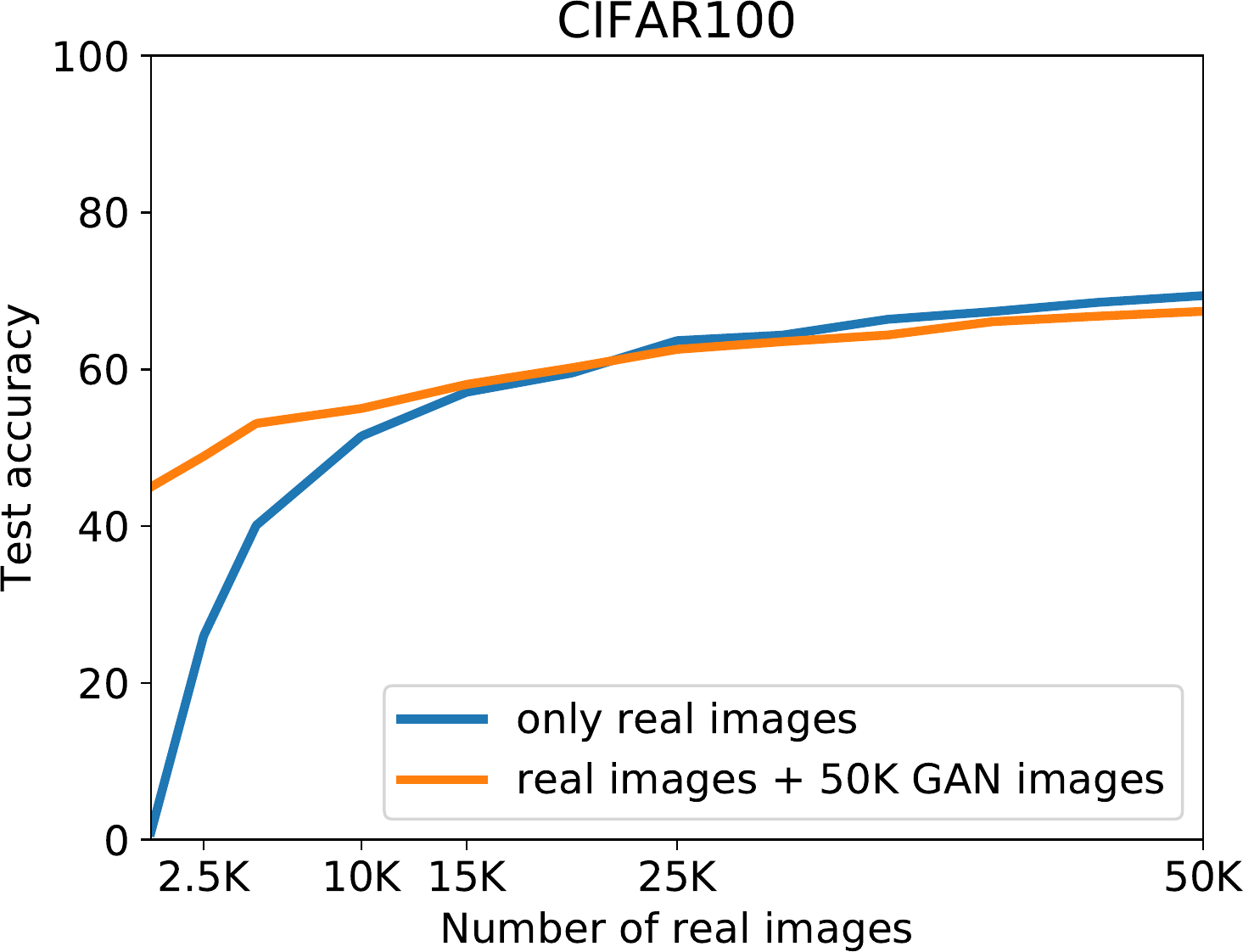} }\label{fig:cifar100_real_data}}%
    \caption{The impact of training a classifier with a combination of real and
      SNGAN generated images.}%
    \label{fig:cifar_real_data}%
\end{figure}

In the case of CIFAR10, we observe that GAN-train accuracy saturates around
15-20k generated images, even for the best model SNGAN (see
Figure~\ref{fig:cifar10_gen_data}). With DCGAN, which is weaker than SNGAN,
GAN-train saturates around 5k images, due to its relatively poorer diversity.
Figure~\ref{fig:cifar100_gen_data} shows no increase in GAN-train accuracy on
CIFAR100 beyond 25k images for all the models. The diversity of 5k
SNGAN-generated images is comparable to the same quantity of real images; see
blue and orange plots in Figure~\ref{fig:cifar100_gen_data}. WGAN-GP (10M) has
very low diversity beyond 5k generated images. WGAN-GP (2.5M) and DCGAN perform
poorly on CIFAR100, and are not competitive with respect to the other methods.

\subsection{GAN data augmentation}\label{sec:augment}
We analyze the utility of GANs for data augmentation, i.e., for generating
additional training samples, with the best-performing GAN model (SNGAN) under
two settings. First, in Figures~\ref{fig:cifar10_real_data} and
\ref{fig:cifar100_real_data}, we show the influence of training the classifier
with a combination of real images from the training set and 50k GAN-generated
images on the CIFAR10 and CIFAR100 datasets respectively. In this case, SNGAN
is trained with all the images from the original training set. From both the
figures, we observe that adding 2.5k or 5k real images to the 50k GAN-generated
images improves the accuracy over the corresponding real-only counterparts.
However, adding 50k real images does not provide any noticeable improvement,
and in fact, reduces the performance slightly in the case of CIFAR100
(Figure~\ref{fig:cifar100_real_data}). This is potentially due to the lack of
image diversity.

\begin{table}[t]
\centering%
\begin{tabular}{l|c|c|c|c}
Num real images & real C10 & real+GAN C10 & real C100 & real+GAN C100\\
  \hline
2.5k & 73.4 & 67.0 & 25.6 & 23.9\\
5k & 80.9 & 77.9 & 40.0 & 33.5\\
10k & 85.8 & 83.5 & 51.5 & 45.5\\
\end{tabular}
\caption{Data augmentation when SNGAN is trained with reduced real image
set. Classifier is trained either on this data (real) or a combination of real
and SNGAN generated images (real+GAN). Performance is shown as \% accuracy.}
\label{tab:data_augm}
\end{table}
This experiment provides another perspective on the diversity of the generated
set, given that the generated images are produced by a GAN learned from the
entire CIFAR10 (or CIFAR100) training dataset. For example, augmenting 2.5k
real images with 50k generated ones results in a better test accuracy than the
model trained only on 5k real images. Thus, we can conclude that the GAN model
generates images that have more diversity than the 2.5k real ones. This is
however, assuming that the generated images are as realistic as the original
data. In practice, the generated images tend to be lacking on the realistic
front, and are more diverse than the real ones. These observations are in
agreement with those from Section~\ref{sec:diversity}, i.e., SNGAN generates
images that are at least as diverse as 5k randomly sampled real images.

In the second setting, SNGAN is trained in a low-data regime. In contrast to
the previous experiment, we train SNGAN on a reduced training set, and then
train the classifier on a combination of this reduced set, and the same number
of generated images. Results in Table~\ref{tab:data_augm} show that on both
CIFAR10 and CIFAR100 (C10 and C100 respectively in the table), the behaviour is
consistent with the whole dataset setting (50k images), i.e., accuracy drops
slightly.

\section{Summary}
This paper presents steps towards addressing the challenging problem of
evaluating and comparing images generated by GANs. To this end, we present new
quantitative measures, GAN-train and GAN-test, which are motivated by precision
and recall scores popularly used in the evaluation of discriminative models. We
evaluate several recent GAN approaches as well as other popular generative
models with these measures. Our extensive experimental analysis demonstrates
that GAN-train and GAN-test not only highlight the difference in performance of
these methods, but are also complementary to existing scores.

\bibliographystyle{splncs04}
\bibliography{egbib}

\clearpage
\section*{Appendix A: Implementation details}
SNGAN in our experiments refers to the model with ResNet architecture, hinge
loss, spectral normalization~\cite{miyato2018spectral}, conditional batch
normalization~\cite{cbn} and conditioning via
projection~\cite{miyato2018cgans}.  We evaluate two variants of WGAN-GP: one
with 2.5M and the other with 10M parameters. Both these variants use ResNet
architecture from~\cite{wgangp}, Wasserstein loss~\cite{wgan} with gradient
penalty~\cite{wgangp} and conditioning via an auxiliary
classifier~\cite{acgan}, with the only difference being that, there are twice
as many filters in each layer of the generator for the 10M model over the 2.5M
variant. We use DCGAN with a simple convnet architecture~\cite{dcgan},
classical GAN loss, conditioned via an auxiliary classifier~\cite{acgan}, as
in~\cite{miyato2018spectral}.

We reimplemented WGAN-GP, SNGAN, DCGAN, and validated our implementations on
CIFAR10 to ensure that they match the published results. Our implementations
are available online~\cite{ourcode}. For PixelCNN++, we used the reference
implementation\footnote{\url{https://github.com/openai/pixel-cnn}} for training
and the accelerated
version\footnote{\url{https://github.com/PrajitR/fast-pixel-cnn}} for
inference.

For all the CIFAR experiments, SNGAN and WGAN-GP are trained for 100k
iterations with a linearly decaying learning rate, starting from 0.0002 to
0, using the Adam optimizer~\cite{kingma2014adam} ($\beta_1 = 0$,
$\beta_2=0.9$, 5 critic steps per generator step, and generator batch size 64).
ACGAN loss scaling coefficients are $1$ and $0.1$ for the critic's and
generator's ACGAN losses respectively. Gradient penalty is 10 as
recommended~\cite{wgangp}. DCGAN is also trained for 100k iterations with
learning rate 0.0002, Adam parameters $\beta_1 = 0.5$, $\beta_2 =
0.999$, and batch size of 100.

In the case of ImageNet, we follow~\cite{miyato2018cgans} for ResNet
architecture and the protocol. We trained for 250k iterations with learning
rate 0.0002, linearly decaying to 0, starting from the 200k-th iteration, for
resolution $64 \times 64$. We use the Adam optimizer~\cite{kingma2014adam}
($\beta_1 = 0$, $\beta_2=0.9$, 5 critic steps per generator step, generator
batch size 64). For $128\times 128$ resolution, we train for 450k iterations,
with learning rate linearly decaying to 0, starting from the 400k-th iteration.
WGAN-GP for ImageNet uses the same ResNet and training schedule as SNGAN. Its
gradient penalty and ACGAN loss coefficients are identical to the CIFAR case.

The classifier for computing GAN-train and GAN-test is a preactivation variant
of ResNet-32 from~\cite{he2016identity} in the CIFAR10 and CIFAR100 evaluation.
Training schedule is identical to the original paper: 64k iterations using
momentum optimizer with learning rate 0.1 dropped to 0.01, after 32k
iterations, and 0.001, after 48k iterations (batch size 128). We also use
standard CIFAR data augmentation: $32\times 32$ crops are randomly sampled from
the padded image (4 pixels on each side), which are flipped horizontally.  The
classifier in the case of ImageNet is ResNet-32 for $64\times 64$ and ResNet-34
for $128\times 128$ with momentum optimizer for 400k iterations, and learning
rate 0.1, dropped to 0.01 after 200k steps (batch size 128). We use a central
crop for both training and testing. We also compute our measures with a random
forest classifier~\cite{randomforests}. Here, we used the scikit-learn
implementation~\cite{scikit-learn} with 100 trees and no depth limitation.

To train SNGAN on MNIST we use the same GAN architecture as in the case of
CIFAR10 (adjusted to a single input channel), and a simple convnet architecture
(four convolutional layers with 32, 64, 128, 256 filters with batch
normalization and max pooling in between, and global average pooling before the
final output layer) as the baseline classifier. The GAN training schedule is
also the same as in the case of CIFAR10. We train the baseline classifier for
64k iterations using a momentum optimizer with learning rate 0.1 dropped to
0.01, after 32k iterations, and 0.001, after 48k iterations (using a batch size
of 128).

\section*{Appendix B: Size comparison}
To complete the discussion on dataset memorization, we compare the size of the
generator and the real dataset. In our CIFAR experiments SNGAN has 8.3M
parameters (33Mb), WGAN-GP has either 2.5M or 10M parameters (10Mb or 40Mb
respectively). In comparison, CIFAR10/100 have 50k training images which
amounts to 150Mb of uncompressed data.

SNGAN for ImageNet has 42M parameters (168Mb) and WGAN-GP has 48M parameters
(192Mb). The entire ImageNet training set takes 16Gb in $64\times 64$
resolution and 64Gb in $128\times 128$ resolution. This difference may
partially explain why compression of CIFAR10/100 into a GAN is relatively
easier than ImageNet.

\section*{Appendix C: t-SNE Embeddings of Datasets}
\begin{figure}[hpb]%
    \centering%
    \subfloat{{\includegraphics[width=0.45\linewidth]{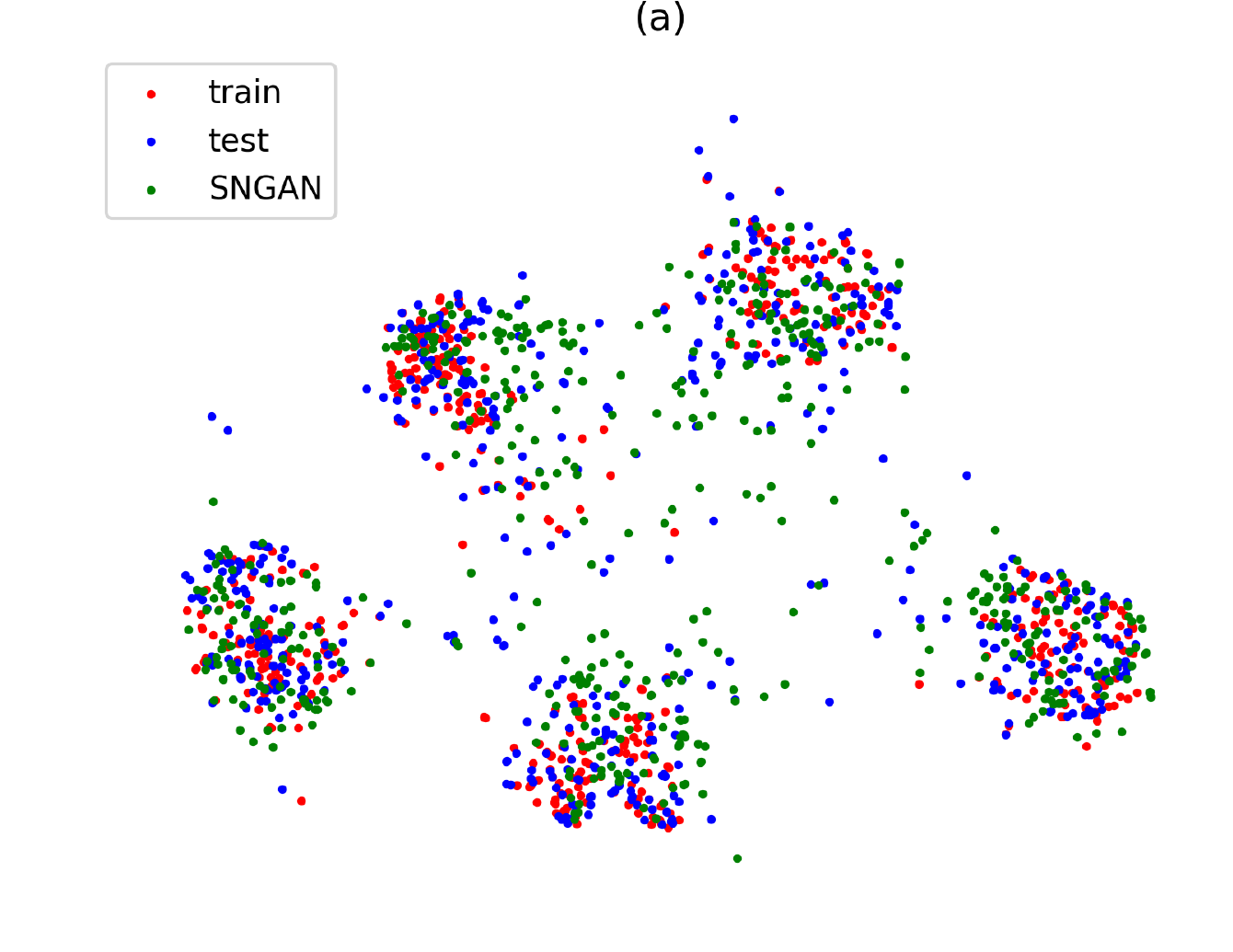} }}%
    \subfloat{{\includegraphics[width=0.45\linewidth]{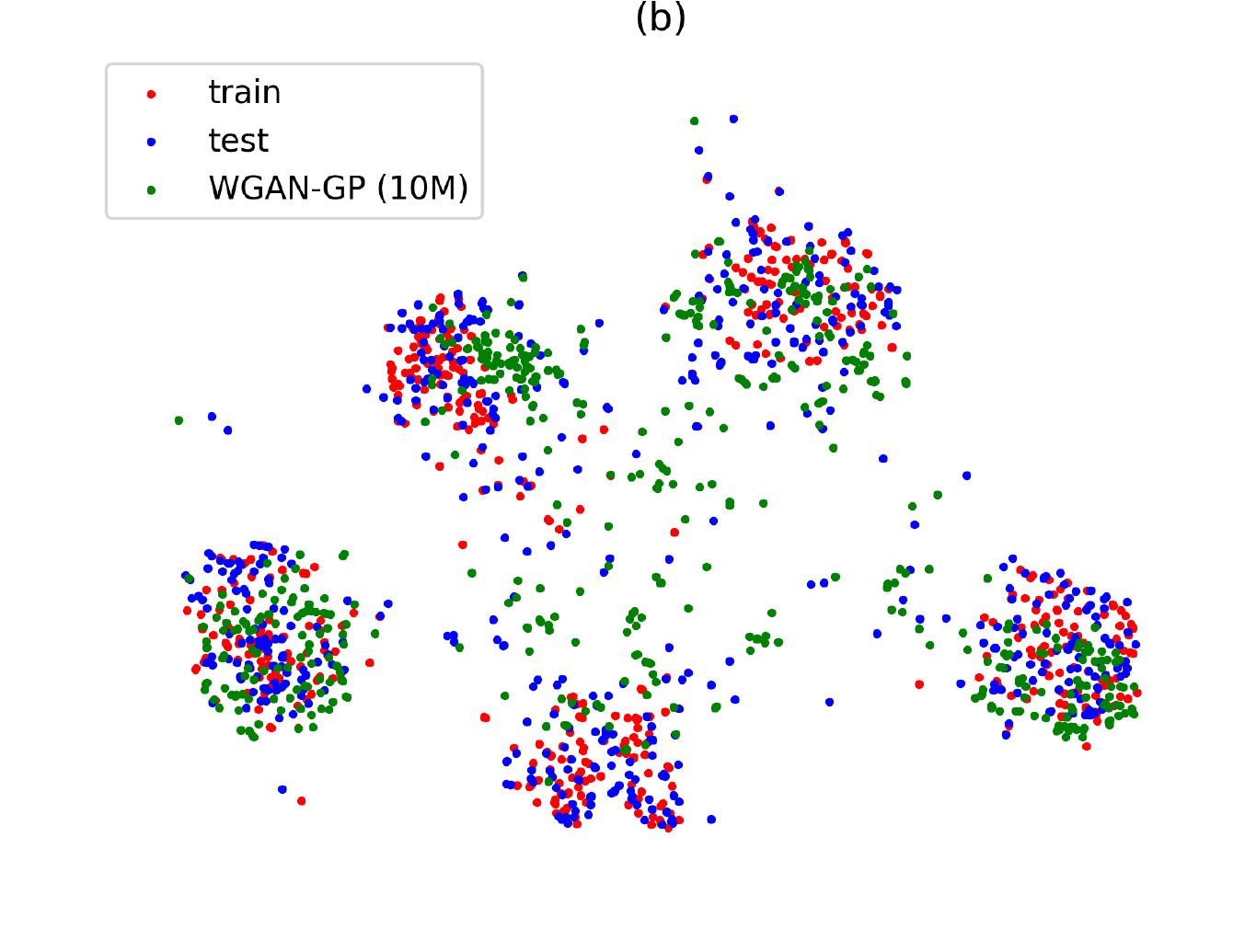} }}%
\\%
    \subfloat{{\includegraphics[width=0.45\linewidth]{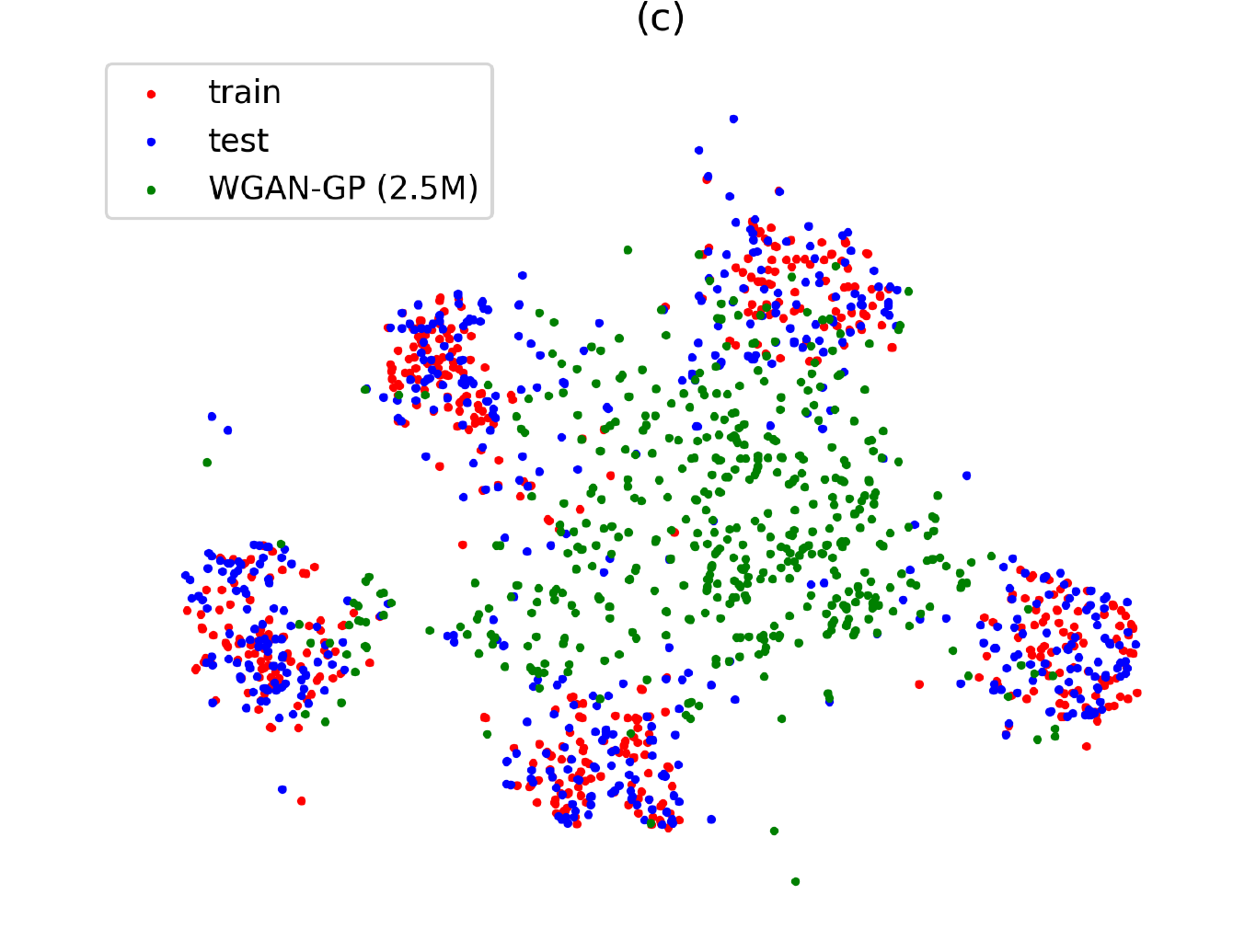} }}%
    \subfloat{{\includegraphics[width=0.45\linewidth]{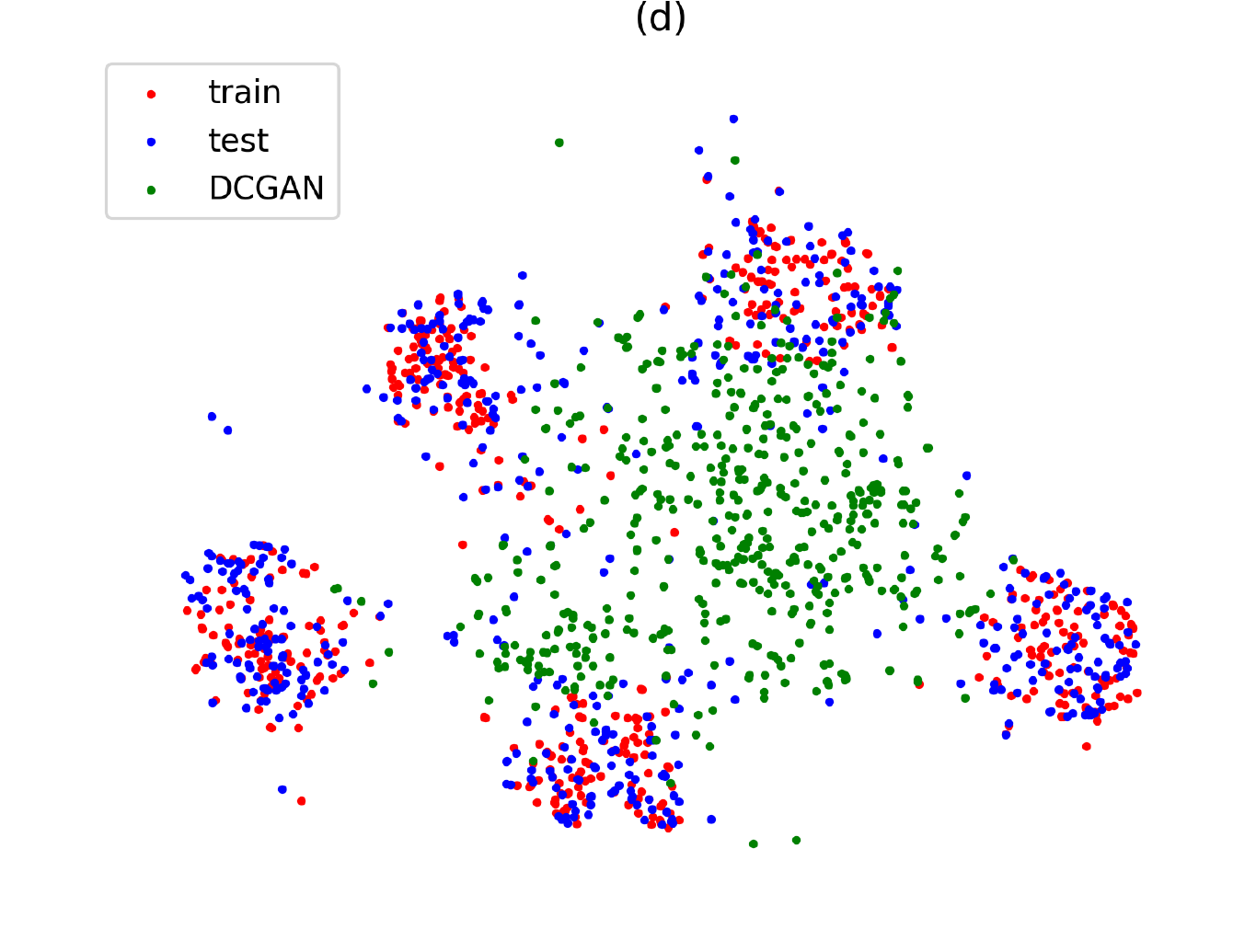} }}%
    \caption{t-SNE~\cite{tsne} embedding of images from 5 CIFAR100 classes,
embedded in the feature space of the baseline CIFAR100 classifier. We use 500
images each from the train and test sets, along with 500 images each generated
with (a) SNGAN, (b) WGAN-GP (10M), (c) WGAN-GP (2.5M), and (d) DCGAN.}%
    \label{fig:tsne}%
\end{figure}

In Figure~\ref{fig:tsne} we show t-SNE embedding of randomly selected images
from the CIFAR100 train and test set splits, and images generated from the four
GAN models (500 images per split or model) of 5 classes (apple, aquarium fish,
baby, bear, bicycle). The t-SNE visualizations are generated with the
embeddings of these images in the feature space of the baseline classifier
trained on CIFAR100, i.e., the classifier used to compute GAN-test accuracy.
Note that the quality of this clustering is in line with the GAN-test accuracy
in Table~\ref{tab:cifar100} in the main paper. For example, the images
generated by SNGAN and WGAN-GP (10M), which produce the two best GAN-test
accuracies, lie close to the training images, while those from WGAN-GP (2.5M)
and DCGAN form a point cloud that does not correspond to any of the clusters.

\end{document}